\documentclass{article} 
\usepackage{nips14submit_e,times}
\usepackage{hyperref}
\usepackage{amsmath}
\usepackage{amsfonts}

\usepackage{array}
\newcolumntype{M}{>{\centering\arraybackslash}m{\dimexpr.25\linewidth-2\tabcolsep}}
\usepackage{pgf}
\usepackage{multirow}
\usepackage{url}
\newcommand\pmf{\mu}
\newcommand\DCM{\textrm{DCM}}
\newcommand\CRP{\textrm{CRP}}
\newcommand\vectorize{\textrm{vec}}

\title{A Bayesian Model of node interaction in networks}

\author{
Ingmar Schuster\\
Natural Language Processing Group \\
Department of Computer Science\\
University of Leipzig\\
\texttt{schuster@informatik.uni-leipzig.de} \\
}

\nipsfinalcopy 

\begin{document}

\maketitle

\begin{abstract}
We are concerned with modeling the strength of 
links in networks by taking into account how often 
those links are used. Link usage is a strong indicator 
of how closely two nodes are related, but existing network
 models in Bayesian Statistics and Machine Learning 
 are able to predict only wether a link exists at all. As priors for latent attributes of network nodes we explore the Chinese Restaurant Process (CRP) and a multivariate Gaussian with fixed dimensionality. The model is applied to a social network dataset and a word coocurrence dataset.
\end{abstract}

\section{Introduction}
\label{intro}

This paper solves two closely related problems. First, given observed counts of interactions between pairs of nodes in a network (input) our models predict the probability of seeing interactions between two nodes with unobserved interaction counts (output). The prediction is achieved  by inferring latent variables for all nodes in the network and a function that maps the latent variables of two nodes to the probability that these nodes interact. Thus, the second problem this paper tackles is uncovering latent structure in network structured data by assigning explicit representations to nodes and inferring an explicit function describing what role node representations play in the probability of an interaction between nodes.

Our models can be readily used for prediction of weighted links in social networks, such as friendship networks or coauthorship networks. Also, they can be used as models of word meaning and the compositionality of meaning in natural languages by identifying each word with a node in a network and word cooccurence counts with interaction counts. Thus, if we know \emph{aggressive} and \emph{dog} frequently appear together in a text corpus, the models should allow to predict, for example, that \emph{aggressive} and \emph{wolf} is a likely combination even if unobserved. For models of word meaning and compositionality, modeling count instead of binary data (\emph{how often} have words cooccurred vs. have they cooccurred \emph{at all}) is essential. If words have cooccurred at all, this does not neccessarily imply they have related meanings. If they cooccur frequently however, the probability that their meanings are related is high. A Bayesian model of word meaning and compositionality of meaning is our ultimate goal, this paper represents a first step.

The contributions of this paper are threefold. First and foremost, since we model interaction counts instead of mere interactions, our models are more expressive than previous work \cite{Miller2009,Palla2012,Lloyd2012}, where the dataset only consisted of adjacency matrices, i.e. matrices containing $1$ if two nodes in a network interact at all, $0$ otherwise. Second, we compare computational aspects and the predictive performance obtained from different priors for the latent representations of nodes. Specifically, we compare the Chinese Restaurant Process ({CRP}) \cite{Aldous1985} and a continuous Gaussian prior of fixed dimensionality. Finally, we refine an existing scheme of sequential intialization for our new likelihod.

We evaluate the properties of our sampler for {CRP} and multivariate Gaussian latent variable models by applying it to the {NIPS} coauthorship dataset and adjective-noun pairs extracted from the Wacky09 corpus \cite{Globerson2007,Baroni2009}.


\section{Model}
We assume both a latent representation $Z_i$ for each node $i$ of the network as well as a latent function $\pmf({\cdot} , {\cdot})$ mapping latent representations $Z_i$ and $Z_j$ (for nodes $i$ and $j$, respectively) to some nonnegative value. Then the marginal probability of seeing an interaction between $i$ and $j$ is
\begin{equation}
P((i,j)) = \frac{\pmf(i,j)}{\sum_{a,b}\pmf(a,b)}
\end{equation}
where $a, b$ range over all network nodes (not accounting for smoothing). We model the function $\pmf(i, j)$ as
\begin{equation}
\pmf(i, j) = \zeta(Z_i^T{ }W{ }Z_j)
\end{equation}
where $W \in \mathbb{R}^{d \times d}$ is a square weight matrix, all latent category representations $Z_a$ have dimensionality $d$ and $\zeta$ is the element wise \emph{softplus}. The \emph{softplus} was introduced as an activation function in Neural Networks \cite{Dugas2001}, is the indefinite integral of the logistic function ${1/1+\exp(-x)}$ and is defined as
\begin{align*}
\zeta: \quad\mathbb{R} &\to \mathbb{R}^{+}\\
x & \mapsto \log(1+\exp(x)).
\end{align*}
It ensures that we only assign positive probability masses. Because the second derivative of the \emph{softplus} is positive, $\zeta$ it is stricly convex, leading to favourabale convergence properties. For a data matrix $C$ containing counts of observed samples $(i,j)$ in row $i$ and column $j$, our likelihood model is
\begin{equation}
\vectorize~C \sim \DCM\left (\vectorize~ \zeta(Z^T{ }W{ }Z  ) \right )
\end{equation}
where $\vectorize~M$ represents vectorization of a matrix $M$. The columns of $Z$ contain latent variables for nodes ($Z = [Z_a Z_b \dots]$). Note the difference in notation from some other papers, where $Z$ contains the latent variables for network nodes in rows instad of columns \cite{Miller2009}. \emph{DCM} is the Dirichlet compound Multinomial
distribution, i.e. a draw from a Dirichlet and a consecutive draw from a multinomial where the Dirichlet draw is integrated out. For a vector $\vectorize~C$ of interaction counts this yields the  probability 
\begin{equation}
p(\vectorize~C\mid\mathbf{\alpha})=\frac{\Gamma\left(A\right)} {\Gamma\left(N+A\right)}\prod_{i,j}\frac{\Gamma(n_{(i,j)}+\alpha_{(i,j)})}{\Gamma(\alpha_{(i,j)})}
\end{equation}
where $\alpha_{(i,j)}$ is the parameter adjusting the prior probability for seeing an interaction between nodes $i$ and $j$, $N$ is the number of draws from the multinomial, $A=\sum_{i,j} \alpha_{(i,j)}$ and $n_{(i,j)}$ is the number of observed interactions ${(i,j)}$.

\subsection{Discrete nonparametric and Continuous parametric Latent Variable Models}
There exists previous work using Gaussian latent variables to represent nodes \cite{Lloyd2012} as well as work using the Indian Buffet Process (IBP) or a hierarchical combination of the {IBP} and the Chinese Restaurant Process (CRP) \cite{Miller2009, Palla2012}. To our knowledge however, no previous paper has compared the convergence and computation time properties of continuous and discrete latent variable spaces for the representation of nodes. We focussed on the comparison of Gaussian (continuous) and CRP (discrete) priors.

The CRP is a nonparametric Bayesian class prior, i.e. it assigns data points (in our case network nodes) to latent classes. The name of the Chinese Restaurant Process stems from the following metaphor: Consider a Chinese restaurant with infinite number of tables. The first customer comes in and sits at the first table. Each of the next customers chooses an already occupied table with probability proportional to the number of customers already sitting there or chooses a new table with some probability. The resulting assignment of customers to tables represents a random draw from the underlying process. Customers  correspond to network nodes in our case, tables act as latent classes. An advantage of the CRP being nonparametric is that the overall number of classes does not need to be chosen as a fix number by the researcher but is inferred from the data. For an introduction, see for example \cite{Aldous1985}.\\
In the case of putting a {CRP} prior on the latent variables representing nodes, the full data model  spells out as follows. The latent variables associated with nodes have a CRP prior
\begin{eqnarray*}
Z^{\CRP} &\sim& \CRP(\alpha_{\CRP})
\end{eqnarray*}
where $\alpha_{\CRP}$ is the CRPs concentration parameter governing the probability of creating a previously unseen latent class. Note that we assume draws from the {CRP} to produce 1-of-K coded column vectors over the latent classes for each node, i.e. the element corresponding to the assigned latent class is $1$, all other components are $0$.

In the case of a multivariate Gaussian prior on latent variables representing nodes we have
\begin{eqnarray*}
Z^{\mathcal{N}}_{i}&\sim&\mathcal{N}(0,\sigma_Z^2~I) 
\end{eqnarray*}
for the latent variable of node $i$. The covariance matrix is diagonal, i.e. there are no dependencies between components of the multivariate Gaussian. Finally, using either of the priors on $Z$, the remaining likelihood model is
\begin{eqnarray*}
W&\sim&\mathcal{N}(W,\sigma_W^2~I)\\
\vectorize~C &\sim& \DCM\left (\vectorize~ \zeta(Z^T{ }W{ }Z  ) \mid \alpha_{\DCM} \right )
\end{eqnarray*}
Here, $W$ is a weight matrix. The covariance matrix for the Gaussian prior on $W$ is again diagonal, mainly for reasons of simplicity but also to avoid problems during slice sampling (see section~\ref{sec:inference}). We put a symmetric prior of $\alpha_{\DCM}/K$ on each component of the {DCM}, where $K$ is the number of seen individual node pairings. The posterior mass for a seen node pair $(i,j)$ then is $\pmf(i,j)+\alpha_{\DCM}/K$. Unseen pairings have a posterior probability of $\sum_{(u,v)~\textrm{unseen}} \pmf(u,v) +\alpha_{\DCM}$. Their prior probability is simply $\alpha_{\DCM}$, which is akin to the usual interpretation of $\alpha_{\DCM}$ being a priori pseudo counts in the Dirichlet distribution.

\subsection{Parallels of Gaussian LV model with Variational Inference approximations}

When approximating the joint posterior of latent variables given the data, two main techniques are prevalent throughout the literature, \textit{sampling} and \textit{variational methods}. In variational methods, often the \textit{mean field assumption} is applied (cf. \cite{Wainwright2007}), i.e. dependencies between latent variables are broken and statistical independence is assumed for the governing variational parameters of the approximating distribution.

The {CRP} is a nonparametric prior on infinite dimensional objects that describes the discrete allocation of network nodes to an unbounded number of classes. However, the flexibility of the normal distribution allows our model to fit the data even with a finite and fixed number of dimensions. We break the model induced dependence of components a priori by using a diagonal covariance matrix for the prior on the latent variables corresponding to nodes.

This leads to a fairly simple and high-performance sampling algorithm that achieves comparable results as when using infinite dimensional nonparametric prior distributions but is much faster.

\section{Related Work}

Our approach is similar to previous approaches which, however, try to predict the probability of a link being present between two nodes $i$ and $j$ rather then the probability of an interaction between them \cite{Miller2009,Palla2012,Lloyd2012}. While \cite{Lloyd2012} notably points out that its approach could be used with likelihoods similar to ours, this is true for other approaches as well \cite{Miller2009,Palla2012}. However, even though the used evaluation datasets would often be naturally modeled as interaction counts, no empirical results have been reported.

In previous Bayesian models, given the probability $l_{i,j}$ of a link between nodes $i$ and $j$, the likelihood of the complete dataset simply is $P(C\mid Z,W) = \prod_{i,j} \left (l_{i,j}^{C_{i,j}} + \left (1-  l_{i,j}\right)^{1-C_{i,j}} \right )$ where $C_{i,j}$ is the component in the data matrix $C$ encoding whether a link exists between $i$ and $j$.

The work by Miller et al.  is closest to ours, the difference being that it uses the {IBP} \cite{Griffiths2005} as the prior on $Z$ while we use the {CRP} or a Gaussian prior, respectively\footnote{We also ran experiments using the IBP as a prior on $Z$. However, due to the different likelihood, for both datasets over $50$ latent dimensions were created and the sampler didn't finish in a week, which is why we cannot report any results.} \cite{Miller2009}. The probability of a link between nodes $i$ and $j$ is $l_{i,j}=f_\sigma(Z_i^T~W~Z_j)$, where $f_\sigma$ is the logistic or probit function.

Palla et al. use a hierarchic combination of the {IBP} and CRP as latent variables representing network nodes, called the \emph{Infinite Latent Attributes (ILA)} model \cite{Palla2012}. Each node is assigned a binary feature vector containing $1$ if a node exhibits a latent attribute and $0$ else. If a node possesses a latent attribute, it might belong to one of several subclusters of the attribute. For subclusters, a {CRP} prior was used. The probability of a link between nodes $i$ and $j$ is $l_{i,j}=f_\sigma(\sum_m Z_{i,m} Z_{j,m} w^{(m)}_{c_i^m,c_j^m} + s)$, where $f_\sigma$ again is the logistic function, $w^{(m)}$ is a weight matrix specifically for the $m$th binary feature, $c_i^m$ is the subcluster assignment for node $i$ in feature $m$, $s$ is a bias term and $m$ ranges over all binary features. In the ILA, only features that are set to $1$ for both nodes influence the likelihood of a link between them, because $Z_{i,m} Z_{j,m}$ is zero if either of the features is zero.

Finally, Lloyd et al. use a Gaussian Process plus logistic function approach to model the probability of a link $l_{i,j}$ from latent variables $Z_i$ and $Z_j$  \cite{Lloyd2012}. They construct a custom kernel function based on the RBF kernel, which ensures that the symmetry properties of a network (i.e. undirected edges) are met. During sampling a multivariate Gaussian with diagonal covariance matrix is used as the prior on $Z$, just like in our approach. Using a Gaussian Process also might be a way to account for interactions between more than two nodes (cf. section~\ref{sec:discussion}). This would be a possibility for modelling multi-word phrases.

\section{Inference}
\label{sec:inference}

We used a Markov Chain Monte Carlo approach to do inference in our model \cite{Andrieu2003}. In the {CRP} case, since we have a finite number of latent classes for finite datasets almost surely, $Z$ can always be stored as a finite matrix. We did not do inference on hyperparameters.

{\bf Sample $Z$ given $W$:}
In every iteration, first the latent variable representation $Z$ for nodes is sampled given $W$. In the case of {CRP} variables, we used Gibbs sampling on $Z$. As Miller et al. point out, the main difficulty arises when computing the likelihood of adding a new class to the representation of $Z$ \cite{Miller2009}. Following their argument, we use a Monte Carlo approximation of the likelihood by repeatedly sampling the weights corresponding to the new class and averaging over likelihoods.\\
In the case of Gaussian variables, each component of $Z$ is slice-sampled with the linear stepping-out procedure \cite{Neal2003}. This simplistic approach is possible because we assumed a diagonal covariance matrix (i.e. zero covariances).

{\bf Sample $W$ given $Z$:} Each component of the weight matrix $W$ can be slice-sampled as well because again we assumed a diagonal covariance matrix.

\subsection{Sequential Initialization}

We found both models to be extremely sensitive to initialization. Here, we adopted and improved a very useful idea for guiding the likelihood to the modes \cite{Palla2012}. Initially two nodes are added to the model and some MCMC iterations are run on the corresponding part of the data. Then all remaining nodes are added in small batches of up to four nodes, running two iterations of MCMC after adding a batch. As expected, initially the sampler is very fast due to the small number of nodes and quickly reaches a high probability region of the parameter space. The latter is because fewer nodes consequently result in fewer local optima.\\
We improved upon this procedure in the following way. In datasets where some nodes interact extremely frequently while other pairs interact rarely, adding new batches can result in very strong changes to the likelihood. We handle this problem by rescaling interaction counts so that the lowest non-zero count is $1$. After all nodes have been added, in each following initialization step the counts of node interactions are multiplied by a constant factor until the original data matrix is recovered.

The problem that this scheme solves does not occur in \cite{Palla2012}, because their task is not factorization of a matrix of counts where elements can range over the non-negative integers, but factorization of an adjacency matrix, where elements are either $0$ or $1$.

\section{Experiments}
\label{sec:experiments}

The {NIPS} dataset \cite{Globerson2007} containing coauthorship information was used in most previous Bayesian network modeling literature by identifying authors with nodes in an undirected graph \cite{Palla2012,Lloyd2012,Miller2009}. An edge was present ($1$) if two authors wrote any non-zero number of papers together, otherwise it was taken as missing ($0$). This is a strong simplification of the coauthorship information in the original dataset.\\
In a social network setting, it might not be of much interest to understand which person knows which other person (this is the task in inferring network edges). A much more interesting question could be how often two persons interact with each other, for example coauthor papers or write messages to each other. This information provides a finer grained measure of how strongly two people relate. Our models account for this setting by identifying people with network nodes and interactions between people (such as sent messages) as count data instead of adjacency in an undirected graph.

We evaluated the models in three ways. First, we report test log likelihood on a held out dataset. Second, we report the rank correlation between true probability of seeing an interaction in the test set and the probability a model assigns to that interaction using Kendall's $\tau$, a common rank correlation coefficient \cite{Kendall1938}. Kendall's $\tau$ takes values in $[-1,1]$, where a correlation of $1$ is interpreted as perfect agreement between two rankings, $-1$ perfect disagreement (i.e. in our case the model prediction ranks would be the reverse of the true probability ranks).
Third, we use the nonlinear \emph{distance corellation} (dcor) measure, which assigns values in the interval $[0,1]$, where a value of $0$ is only assigned in the case of statistical independence \cite{Szekely2007a,Kosorok2009}.

\subsection{Types of held out data}
There are two sensible ways of holding out data depending on the problem. For each scheme, we used $80\%$ of a dataset for training, $20\%$ for testing.

\begin{figure}[p]
\begin{center}
\begin{tabular}{cc}
Interactions held out & Node pairs held out \\
\includegraphics[width=0.49\columnwidth]{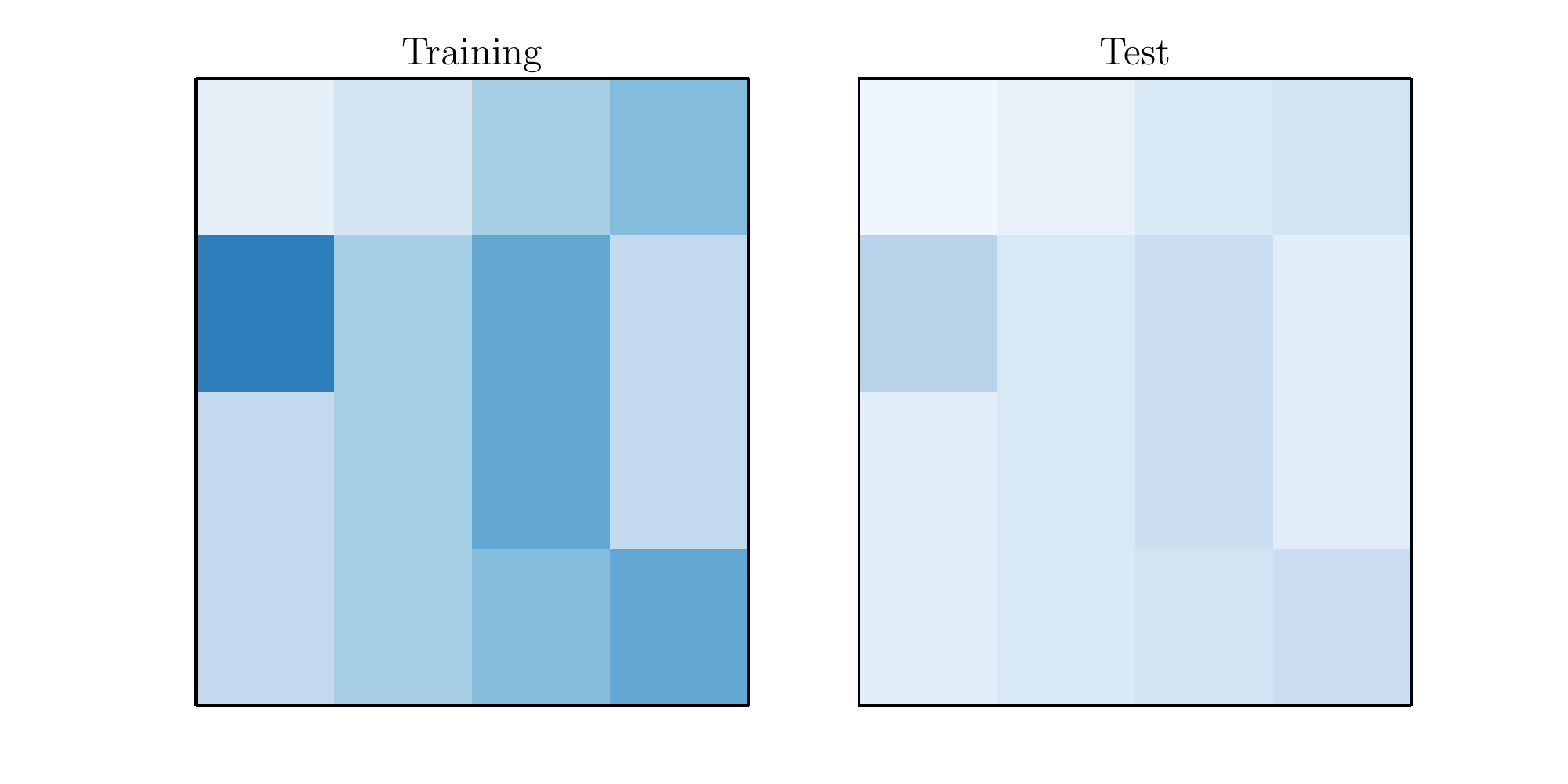} & \includegraphics[width=0.49\columnwidth]{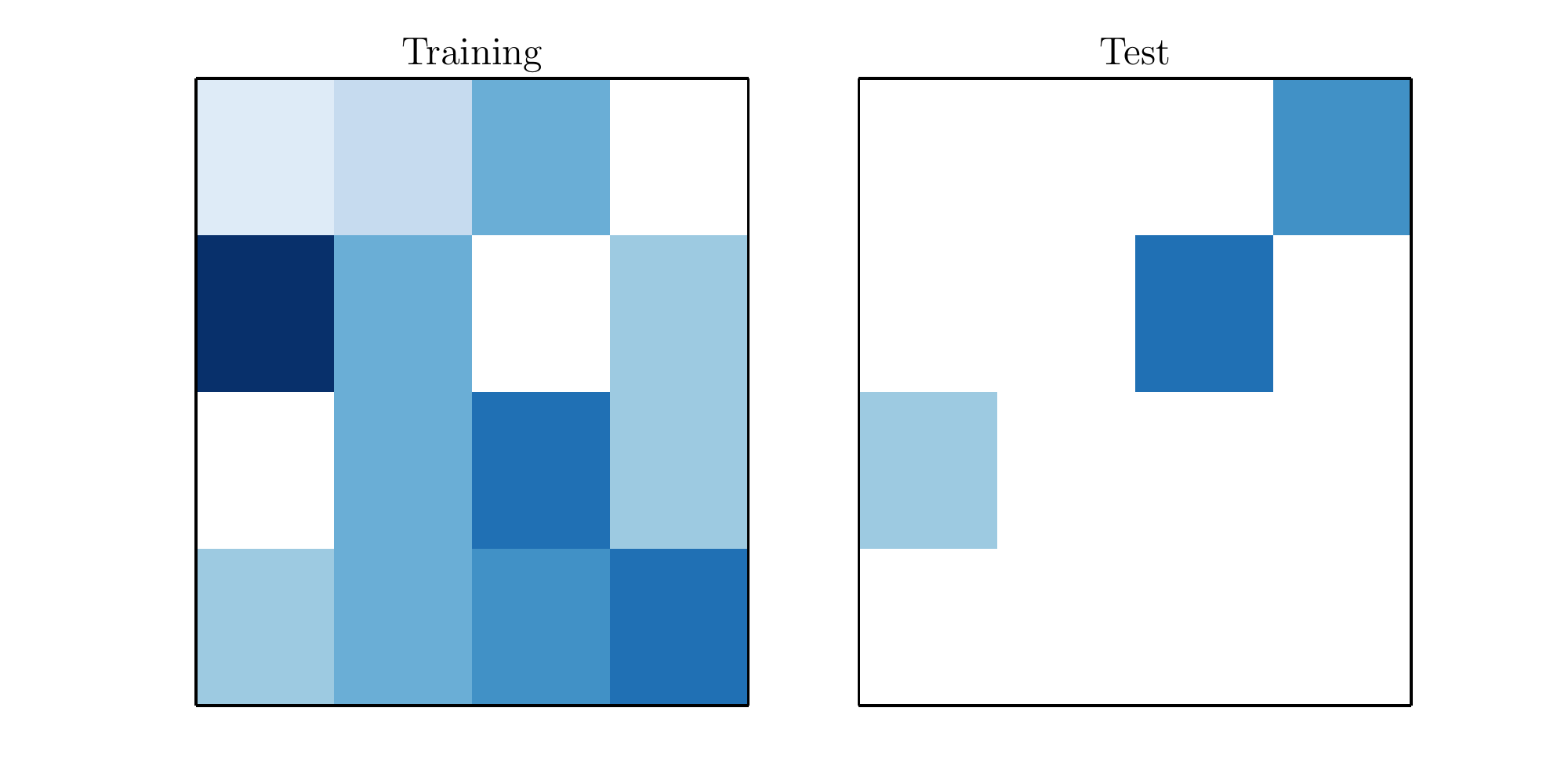}

\end{tabular}
\caption{Holding out interactions versus holding out node pairs}
\label{fig:ex_ho_obs}
\end{center}
\end{figure}

{\bf Held out interactions} When using the model to predict future interactions or just to build a more compact representation of the probability of interactions, test data might consist of some observations of interactions withheld from the count matrix (Figure \ref{fig:ex_ho_obs} left). For example if a dataset contains $10$ cooccurrences of the words \emph{aggressive dog}, $8$ of these go into the training data, $2$ into the test data.

{\bf Held out node pairs} All interactions for some set $H$ of node pairs are held out. The training dataset does not contain counts for these, the test dataset contains only counts for these (treat as missing data, Figure \ref{fig:ex_ho_obs} right). Obviously, this is the more difficult of the two schemes of holding out data. However it is also the more interesting scheme, as it can be interpreted as the transfer of knowledge from interactions between one part of the nodes to another, unknown part.

\begin{figure}[p]
\begin{center}
\includegraphics[width=\textwidth]{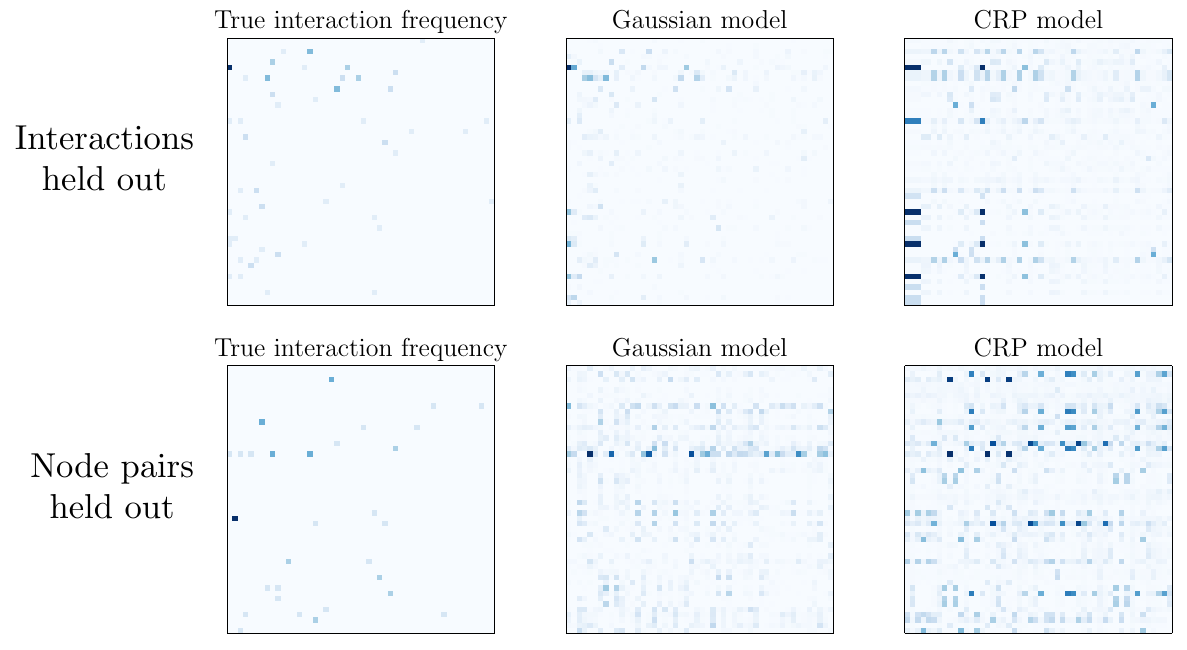}
\caption{{NIPS} dataset. Darker means higher probability, nodes subsampled for clarity.}
\label{fig:nips_matrix}
\end{center}
\end{figure} 

\begin{figure}[p]
\begin{center}
\includegraphics[width=\textwidth]{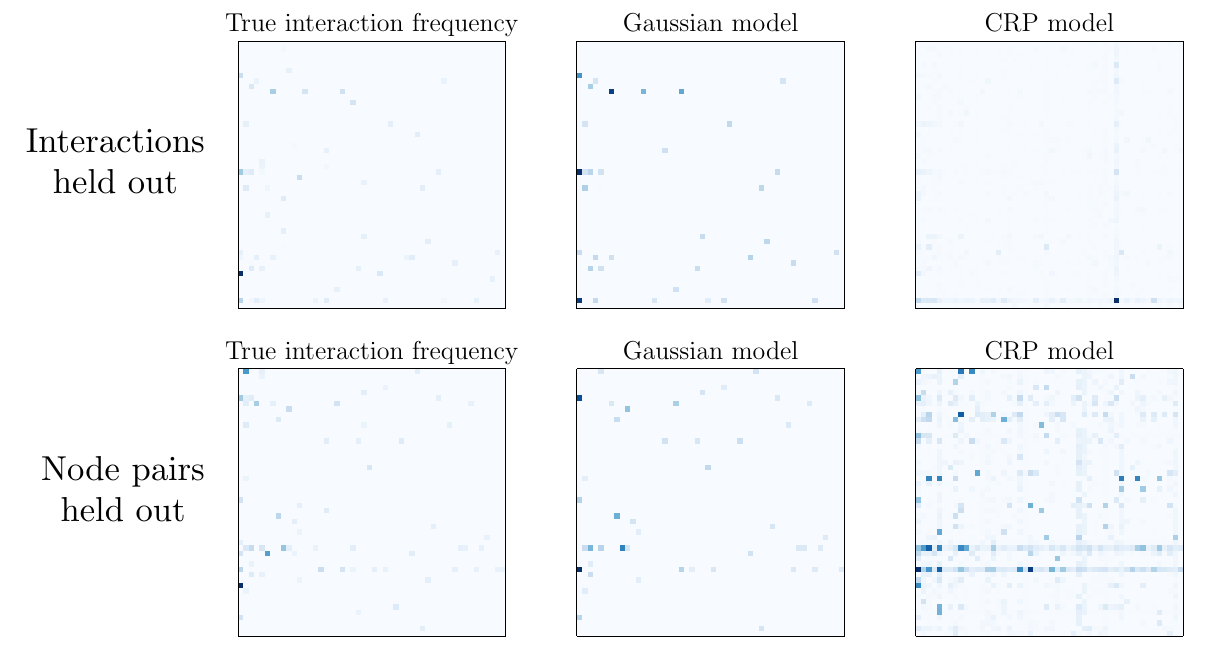}
\caption{{WaCky} dataset. Darker means higher probability, nodes subsampled for clarity.}
\label{fig:wacky_matrix}
\end{center}
\end{figure} 

\subsection{{NIPS} coauthorship dataset}
\label{sec:nips}
The NIPS dataset contains a list of all papers and authors from the NIPS conferences 1-17, enabling the extraction of the number  of interactions between two authors (number of papers coauthored)  \cite{Globerson2007}. Like in most previous work employing this data set, we used the 234 authors who had published with the most other people, resulting in $528$ overall interactions \cite{Miller2009,Palla2012,Lloyd2012}. However, instead of reducing the dataset to the information whether or not two authors had written a paper together (binary values), we constructed a count matrix from the number of papers a pair of two authors $i$ and $j$ had published together.

\begin{figure}[ht]

\begin{tabular}{cc}
Interactions held out & Node pairs held out \\
\includegraphics[width=.48\textwidth]{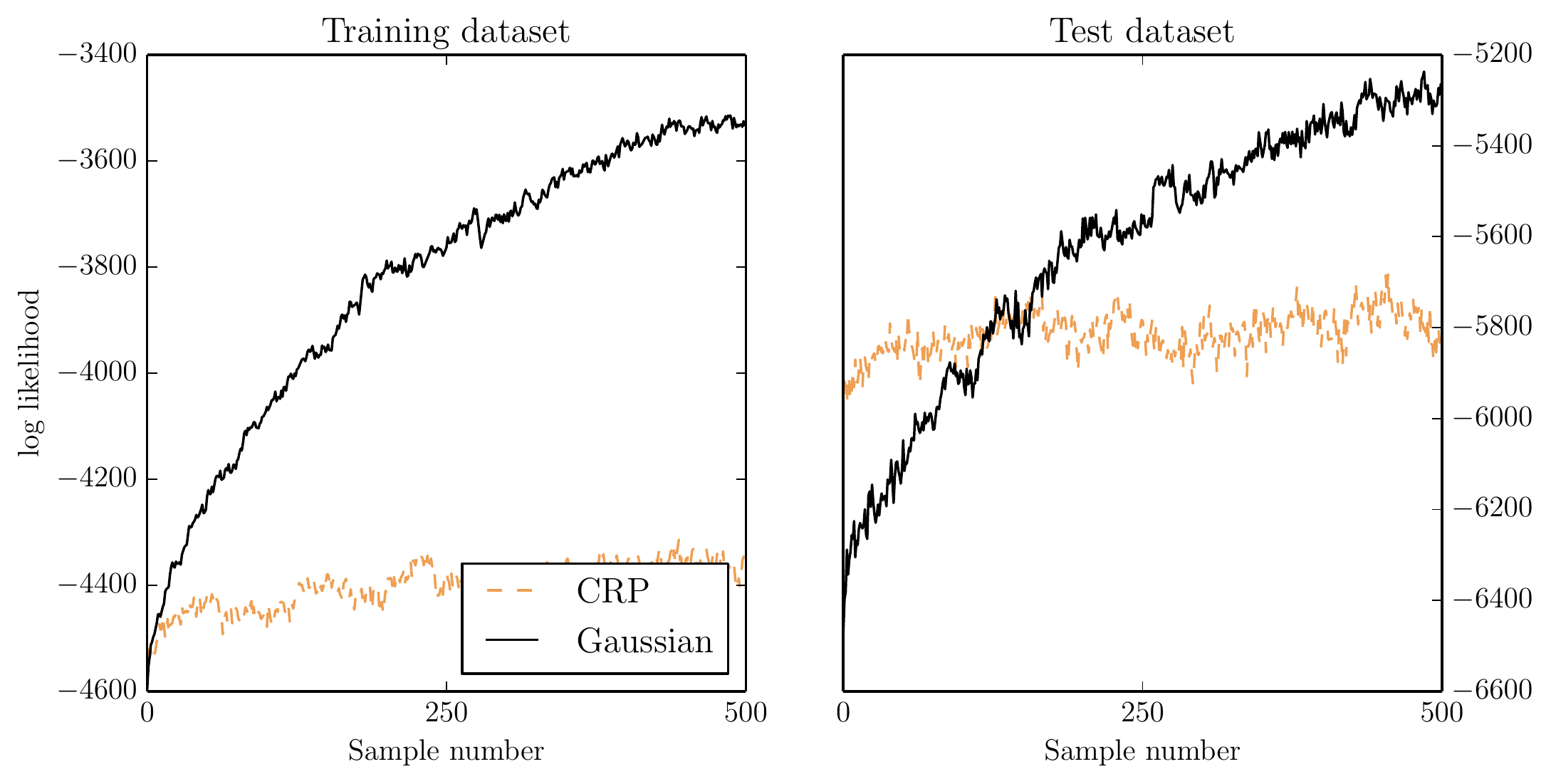}
&
\includegraphics[width=.48\textwidth]{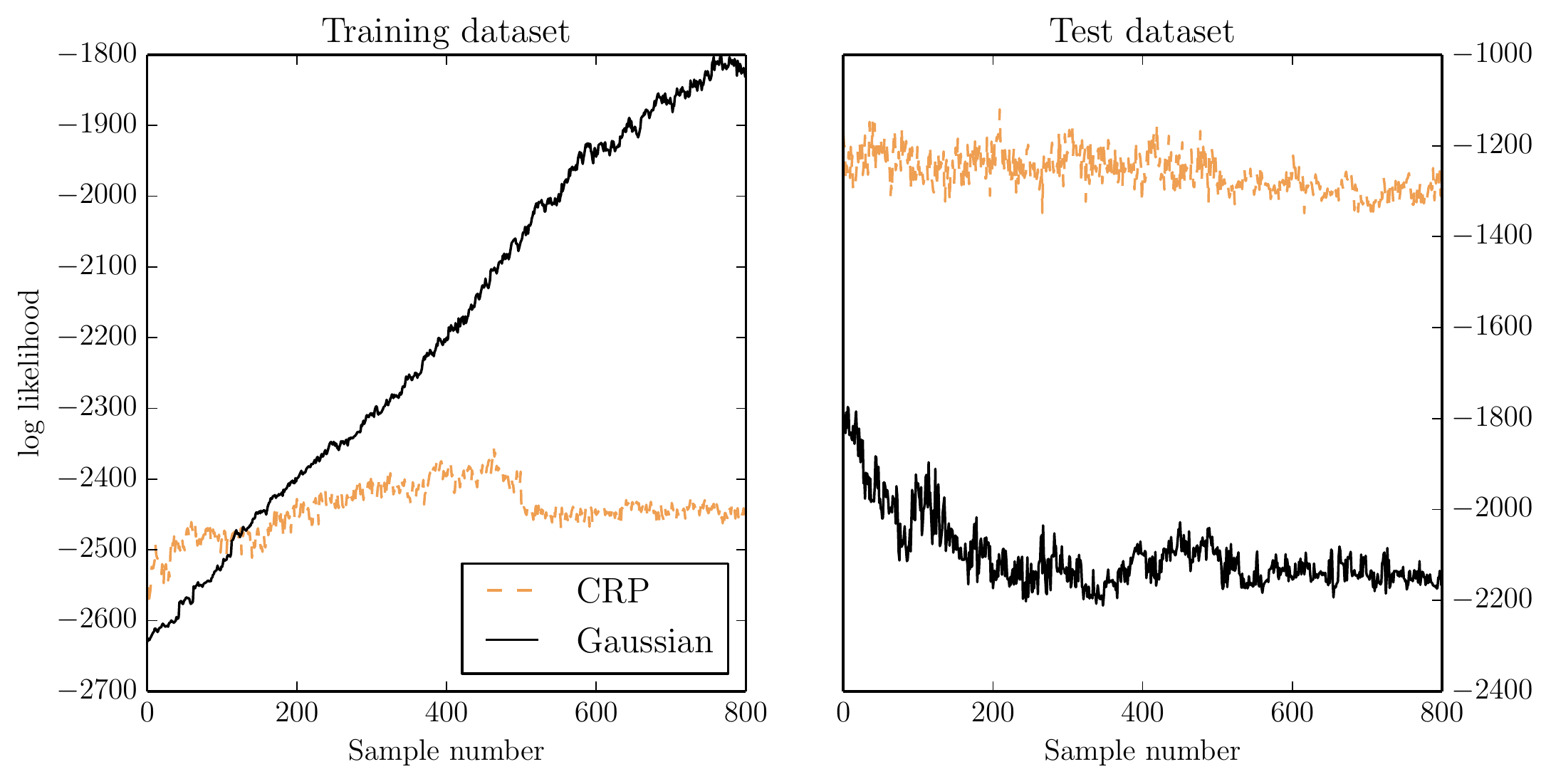}
\end{tabular}

\caption{{NIPS} data trace plots}
\label{fig:nips_trace}

\end{figure}

We collected $500$ samples for this dataset; for the node pairs held out scheme we decided to collect an additional $300$ samples to ensure convergence of the algorithm. The Gaussian model was very fast to mix in the held out interactions scheme, see Figure~\ref{fig:nips_trace} (left). This is no surprise when considering that for the Gaussian model we used slice sampling of each component of a latent variable, whereas in the {CRP} model we had to fall back to Gibbs sampling. Another aspect here is that computation time per sample was much better for the Gaussian model (see Table \ref{tab:stat_nips}). However, a possible reason for this might be that in the {CRP} model, for each node the likelihood of belonging to a previously unseen latent class had to be estimated using Monte Carlo integration over the corresponding entries in $W$. This was not the case for the Gaussian model which had fixed dimensionality.

Kendall's $\tau$ showed better rank correlation for the Gaussian model in both schemes of held-out data. However, correlation was not very high in the held out pairs scheme. One possible cause for this is the extreme data scarcity. The dataset contains only $528$ interactions for $234^2$ node pairs. Data scarcity might also be the reason for the difference between dcor and $\tau$ measures in the held-out node pairs scheme. For the comparatively large {WaCky} dataset discussed in section \ref{sec:wacky} there is no disagreement between these measures.




\begin{table*}[tb]
\caption{Statistics of inference on the {NIPS} dataset}
\label{tab:stat_nips}
\vskip 0.15in
\begin{center}
\begin{tabular}{ccccccccc}
\hline
{Held out type} & Model & Dimens. & sec/sample & Kendall's $\tau$ (\emph{p} value)& dcor & test ll  \\

\hline
\multirow{2}{*}{Interactions} & {CRP}  & $21.7$ & $996$ & $0.3559~(7~e-24)$ & 0.3837& $-5625.08$ \\
							  &  Gaussian	& $6$ & $428$ & $0.4479~(8~e-37)$ & $0.5685$& $-5625.08$ \\

\multirow{2}{*}{Pairings}& {CRP}  & $15.2$ & $581$ & $0.0169~(0.8554)$ & $0.2661$ & $-2100.53$ \\
					& Gaussian	& $6$ & $302$ & $0.0710~(0.4441)$ & $0.1395$ & $-2100.53$ \\
                               

\hline                  
\end{tabular}
\end{center}
\vskip -0.1in
\end{table*}

The {CRP} model shows more artifacts than the Gaussian model in Figure \ref{fig:nips_matrix}. Especially interesting is the difference in training and test log likelihood for this dataset (Figure \ref{fig:nips_trace}). The training log likelihood of the Gaussian model quickly outperforms that of the {CRP} model in both schemes of held out data. Still, for held out node pairs, the test log likelihood is better for the {CRP} model in every single sample (see Figure \ref{fig:nips_trace}, right). Since the {CRP} model, as opposed to the Gaussian model, seems to gravitate towards a uniform distribution, it can closely approximate the little data available. The situation is slightly different for the next dataset.

\subsection{WaCky adjective-noun coocurrence dataset}
\label{sec:wacky}
This dataset was comprised of adjective-noun word pairs extracted from the {WaCky09} corpus \cite{Baroni2009}. Every word was identified with a network node, while cooccurences of words where identified with node interaction (e.g. \emph{aggressive dog} was counted as an interaction between the \emph{aggressive} and \emph{dog} nodes). We used $40000$ sentences, resulting in a total of $210$ words and $22582$ interactions (cooccurences).

We collected $1500$ samples. Correlation between true and model probability of seeing a node interaction were better for the Gaussian model than for the {CRP} model in both held-out schemes (see Table \ref{tab:stat_wacky}). In the held-out node pairs scheme, the rank corellation between true and model probability was even negative for the {CRP} model. However, rank correlation did not reach statistical significance in the held-out node pairs scheme. Again, computation time seems to favor the Gaussian model, although further experiments would be needed with fewer dimensions in the held-out node pairs scheme\footnote{The fact that a sample was faster as compared to a NIPS dataset sample (and the fact that we collected many more samples) stems from caching a lot of intermediate variables, saving computation time}.

\begin{figure}[ht]

\begin{tabular}{cc}
Interactions held out & Node pairs held out \\
\includegraphics[width=.48\textwidth]{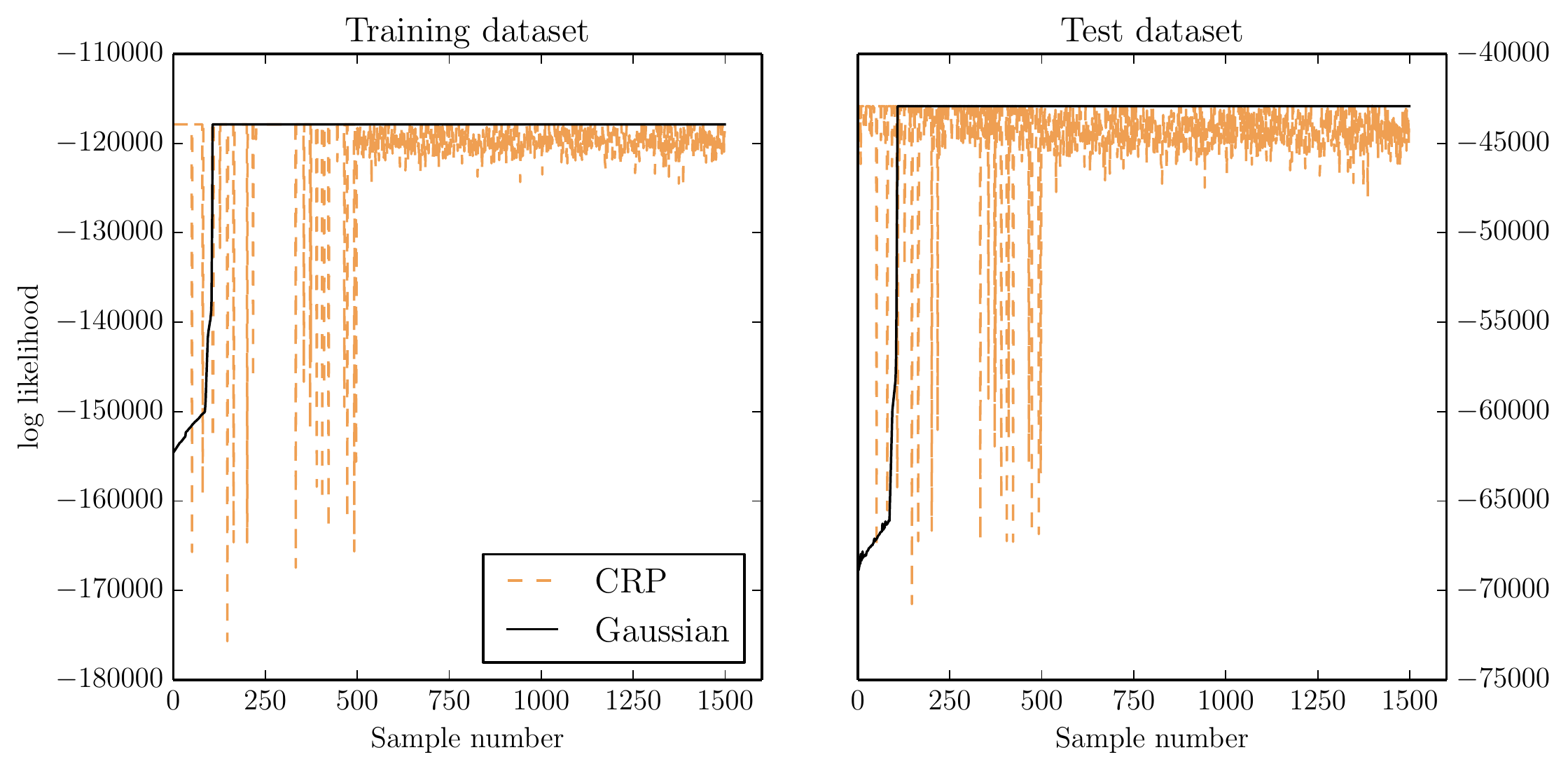}
&
\includegraphics[width=.48\textwidth]{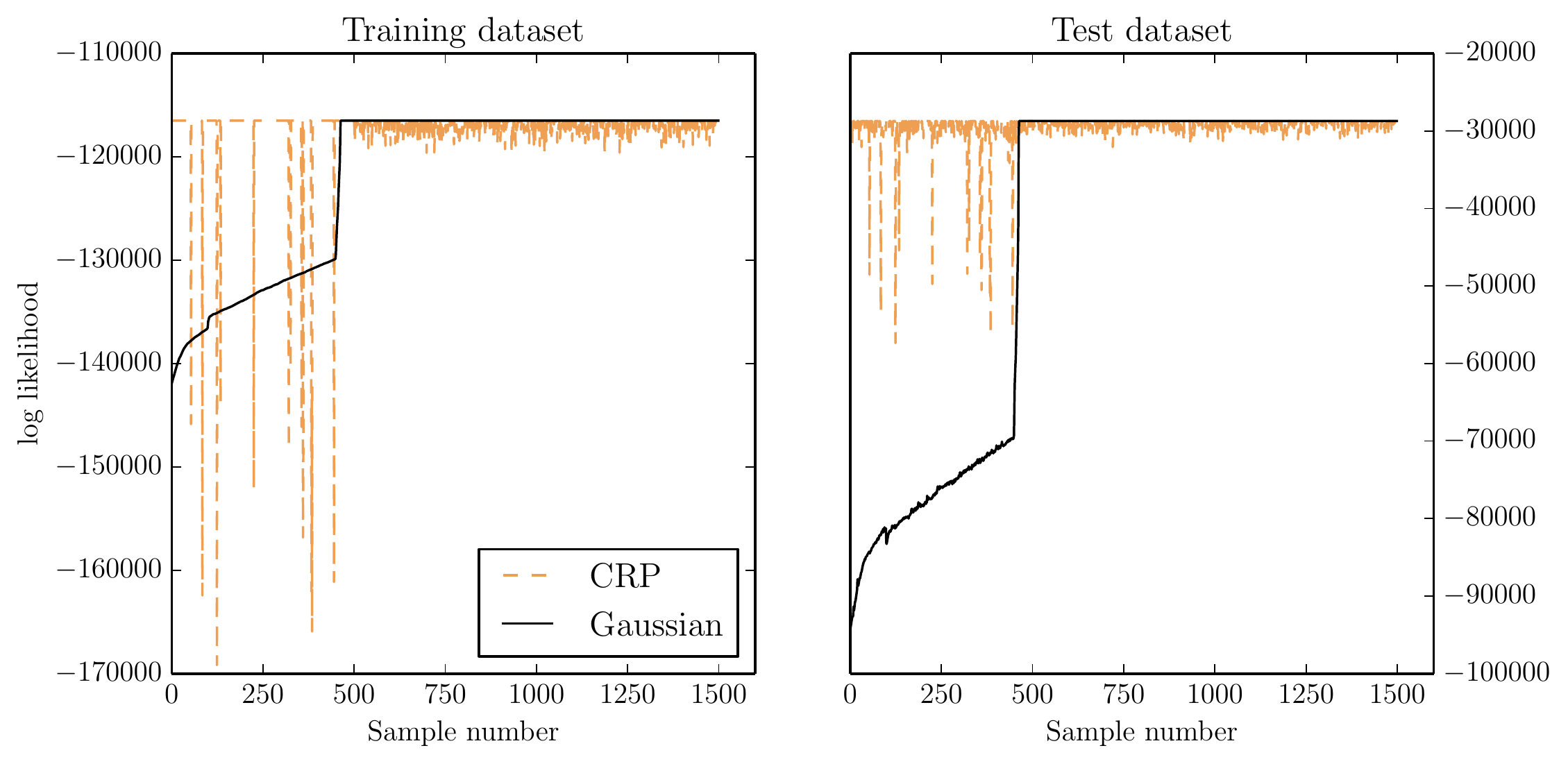}
\end{tabular}

\caption{{WaCky} data trace plots}
\label{fig:wacky_trace}

\end{figure}

Like in the {NIPS} dataset there are clear artifacts for the {CRP} model (see Figure \ref{fig:wacky_matrix}). The trace plot for this larger data set (Figure \ref{fig:wacky_trace}) suggests that the likelihood modes are extremely peaky.

\begin{table*}[tb]
\caption{Statistics of inference on the {WaCky} dataset}
\label{tab:stat_wacky}
\vskip 0.15in
\begin{center}
\begin{tabular}{ccccccccc}
\hline
{Held out type} & Model & Dimens. & sec/sample & Kendall's $\tau$ (\emph{p} value)& dcor & test ll \\

\hline
\multirow{2}{*}{Interactions} & {CRP}  & $5.7$ & $234$ & $0.2358~(1~e-9)$ & $0.2897$& $-44565$ \\
							  &  Gaussian	& $8$ & $155$ & $0.6266~(9~e-58)$ &$0.6561$& $-44565$ \\



\multirow{2}{*}{Pairings}& {CRP}  & $4.5$ & $123$ & $-0.1137~(0.2075)$ & $0.2164$ & $ -43642$ \\
						 & Gaussian	& $8$ & $175$ & $0.1604~(0.0753)$ & $0.3033$ & $ -43642$ \\


                                
\hline                  
\end{tabular}
\end{center}
\vskip -0.1in
\end{table*}

\section{Discussion and Future Work}
\label{sec:discussion}

In this paper, we have explored the problem of inferring latent structure from counts of interactions between nodes in a network. Our model factorizes the interaction count matrix into a latent representation for network nodes and a weight matrix that encodes a PMF. The prior that has been used for latent node representations is either the Chinese Restaurant Process (CRP) or a multivariate Gaussian with fixed dimensionality. We have evaluated our models on the {NIPS} coauthorship dataset and an adjective-noun word pair dataset extracted from the {WaCky09} corpus \cite{Globerson2007,Baroni2009}.

The paper has made two main contributions. Most importantly, we modeled interactions between network nodes instead of the mere existence of a link between nodes. Second, we compared computational and predictive properties of a discrete and a continuous prior for latent variables associated with network nodes, finding the latter advantagous over the first, both in computation time and ability to fit the data. Another contribution was the refinement of a sequential initialization scheme for our new likelihood.

Future work should enable inference in our Gaussion model in the streaming data setting. One possibility would be to use a sampling scheme using stochastic gradient langevin dynamics with artificial noise with our Gaussian model \cite{Welling2011}. However, a stochastic gradient descent optimization solution might not be much worse than using sampling, given that the data likelihoods seem to be very peaky. \\
Another idea worth pursuing is to remove the fixed dimensionality constraint of our Gaussian model. One viable and simple way would be to put a prior on the number of dimensions, like done for the alternative of the Dirichlet Process developed by Miller \& Harrison \cite{Miller2013}. This approach has the advantage of not overestimating the number of components.

As has been suggested, we also aim to extend our models to the case of interactions between multiple nodes to account for multi-word phrases (as opposed to just pairs of words). This could be achieved by replacing the linear function represented by the weight matrix $W$ by a Gaussian Process with a kernel based on a matrix-norm induced distance measure between the latent representations of network nodes. Such a model would also allow to account for more than two coauthors in the NIPS dataset.

\section*{Acknowledgments} 
 I thank Patrick J\"ahnichen and anonymous reviewers for helpful comments and proofreading.

\bibliography{library}
\bibliographystyle{unsrt}

\end{document}